# ArtHDR-Net: Perceptually Realistic and Accurate HDR Content Creation

Hrishav Bakul Barua*[†], Ganesh Krishnasamy*, KokSheik Wong*, Kalin Stefanov[‡], and Abhinav Dhall[‡§]
* School of Information Technology, Monash University, Malaysia
[†] Robotics and Autonomous Systems Group, TCS Research, India
[‡] Faculty of Information Technology, Monash University, Australia
[§] Centre for Applied Research in Data Sciences, Indian Institute of Technology Ropar, India

*Abstract*—High Dynamic Range (HDR) content creation has become an important topic for modern media and entertainment sectors, gaming and Augmented/Virtual Reality industries. Many methods have been proposed to recreate the HDR counterparts of input Low Dynamic Range (LDR) images/videos given a single exposure or multi-exposure LDRs. The state-of-the-art methods focus primarily on the preservation of the reconstruction's structural similarity and the pixel-wise accuracy. However, these conventional approaches do not emphasize preserving the artistic intent of the images in terms of human visual perception, which is an essential element in media, entertainment and gaming. In this paper, we attempt to study and fill this gap. We propose an architecture called *ArtHDR-Net* based on a Convolutional Neural Network that uses multi-exposed LDR features as input. Experimental results show that *ArtHDR-Net* can achieve state-of-the-art performance in terms of the HDR-VDP-2 score (i.e., mean opinion score index) while reaching competitive performance in terms of PSNR and SSIM.

## I. INTRODUCTION

The High Dynamic Range (HDR) imaging is introduced to handle the scenario where the visual content has a high ratio between the brightest and the darkest pixels [1]. HDR allows the inclusion of multiple dynamic exposures in a single visual content. Hence, if an image is in HDR, it has a wide gamut of brightness levels, i.e., there is a significant variation in light levels within a scene with high details. In other words, the well-lid and dark regions in a scene are captured with equal clarity using HDR technology. The main goal of HDR imaging is to capture the real world lighting, greater luminance distribution, brightness, detailed shadows, and radiance in a manner which is as realistic as possible and can mimic the Human Visual System and its level of perception. HDR technology has many applications, e.g., in the fields of media and entertainment [1], Augmented/Virtual Reality (AR/VR) [2], [3], gaming and visualization [2], [4], and computational photography [5]. With the increase in HDR supported hardware devices such as HDR displays, AR/VR sets, mobile screens, and cameras, researchers have tried to find ways and methods to successfully retrieve the HDR counterparts from the conventional Low Dynamic Range (LDR) content, known as inverse tone mapping.

HDR content can be created using special hardware (e.g., expensive cameras and visual sensors) and/or software (e.g., virtual environments and renderers). However, converting the existing LDR content to HDR requires unique methods and algorithms [6], [7] to expand the color gamut and dynamic range, which is usually achieved through inverse tone mapping and image enhancement. In this paper, we attempt to take an LDR image as input and create an HDR image with perceptually realistic and visually accurate information in terms of the Human Visual System represented by the HDR-VDP-2 scores (see section IV-B). This work contrasts most state-of-the-art methods that attempt to solve the problem of structural similarity of the generated HDR with respect to the LDR input. The main contributions of this work are threefold:

- We propose a CNN-based network (Section III) for HDR image content creation drawing inspiration from the method in [8]. We leverage the features from three different exposure values (EV -2, 0, and +2) and create a combined feature map for a better representation of under/over-exposed areas in the LDRs.
- We perform a thorough experimental validation with quantitative and qualitative comparisons (Sections IV and V) to establish the efficacy of our contributions.
- We present a detailed ablation study (Section VI) on the different losses used to train the network and the skip connections of the CNN layers utilized for better representation of the low-level features.

## II. RELATED WORK

Researchers have employed various Machine Learning (ML) and Deep Learning (DL) methods [7] for HDR content creation. Below, we categorize the recent relevant literature into two major buckets, i.e., image- and video-based. These buckets can be further categorized based on single and multiple exposure LDR inputs.

### A. Image-based Methods

Khan et al. [8] proposed a feedback-based Convolutional Neural Network (CNN) to create an HDR image from a single exposure LDR input image. Their method uses a dense feedback mechanism based on Recurrent Neural Network (RNN) on top of a conventional CNN to obtain both low-level and high-level features. This coarse-to-fine feature representation in various iterations leads to HDR creation at a higher quality. Phuoc-Hieu Le et al. [12] proposed a Neural Network (NN) architecture to learn and generate multiple exposures from a single image. The model can reconstruct pixel radiance and details of hallucination in different exposures by inverting the camera response. Yu-Lun Liu et al. [9] proposed a CNN architecture that learns the reverse process of camera pipeline, i.e., inverse tone mapping. The method employs three CNNs to reverse the three basic steps in camera pipeline, i.e., clipping of dynamic range, camera response function (CRF) based non-linear mapping, and quantization. Then the three sub-tasks are jointly fine-tuned to reduce the accumulated errors. Yet another interesting work [13] (although not in the DL domain) talks about an inverse tone mapping operator, which can preserve the artistic intent of the content creator in the generated HDR images i.e., the images look more realistic perceptually. Jinghui Li et al. [14] presented a novel multi-scale CNN with an attention mechanism that tackles the problems of color quantization and the loss of information in over/under-exposed regions. The method re-scales channel-wise features using a channel attention method adaptively. Another DL based method [15] proposed a novel architecture that can recover the saturated pixels of the LDR images with better quality leading to visually and perceptually better HDR images with well-preserved textures. The method employs a feature masking mechanism that limits the contribution of the features from the saturated regions of the images, which in turn reduces the checkerboard and halo artifacts of these regions. Gabriel Eilertsen et al. [16] proposed a CNN-based method that overcomes the challenges of predicting HDR pixels from complex LDR images with over/under exposed areas. Another interesting work [17] considered the low peak brightness issues and the missing of artistic intent of the existing methods. It proposed a mid-level automatic tone-mapping operator. Gaofeng Cau et al. [18] proposed a spatial and channel dimensional decoupled kernel prediction network that combines a simple module which produces preliminary HDR results and an encoder/decoder module to produce pixel-wise kernels. Finally, these two results are convolved to produce high-quality HDR images. The authors claim that the method can use the relevant information from under- and over-exposed images while avoiding pixel-wise degradation and information redundancy across channels. A work by Hanbyol Jang et al. [19] porposed a novel method of HDR generation from LDR based on histogram and color difference between HDR-LDR (multi-exposure) pairs.

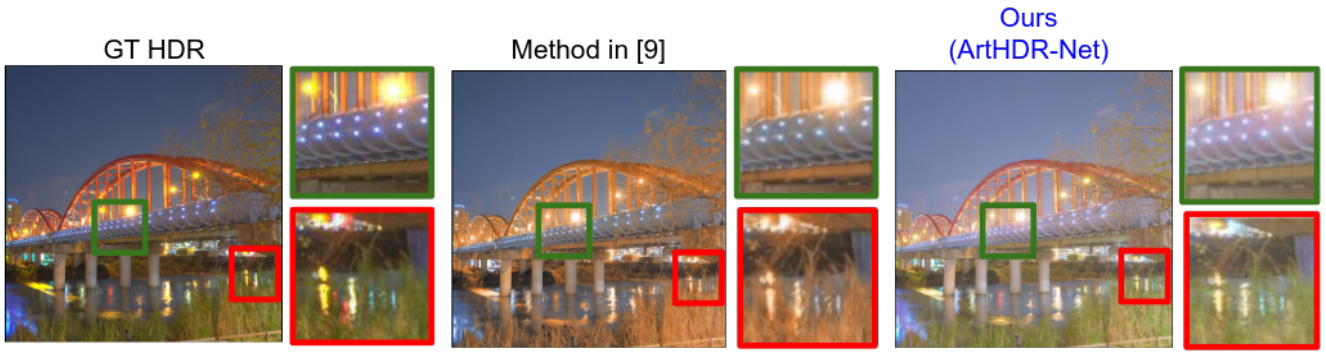

Fig. 1: Output produced by our method ArtHDR-Net compared to one of the closest state-of-the-art [9] methods in terms of HDR-VDP-2 score [10], [11]. Our method reconstructs the HDR content with better naturalness in terms of color contrast and luminance levels.

### B. Video-based Methods

Chen et al. [1] proposed a new architecture for reconstructing HDR videos from LDR counterparts based on a divide-and-conquer policy. The final framework has three components: adaptive global color mapping, local enhancement, and highlight generation. Yang et al. [20] proposed an interesting framework which is capable of utilizing two modalities of input, i.e., stacked events and the LDR frames to create a learnt latent space. The method is guided by the events to facilitate multi-modal learning for HDR creation. The method decribed in [21] proposed a Generative Adversarial Network (GAN) approach for reconstructing HDR videos from multiple exposed LDR sequences. The method starts with a denoising module to clean the LDR input and uses optical flow to align the neighbouring multi-exposed LDRs in the sequence.

### C. Positioning Our Method

Almost all of the discussed state-of-the-art methods focus on improving the overall structural quality of the generated HDR images while preserving the original structure in the LDR inputs. However, only a handful (i.e., less than five) of the works try to tackle the problem of generating HDR images from the perspective of the Human Visual System and creating perceptually pleasing image for the human eye (see Figure 1). Most conventional methods also lack discussion of the preservation of the artistic intent [13] of the generated HDR content. In this work, we attempt to address these limitations in the current literature along with structural similarity preservation. We achieve better peak brightness with a deep CNN-based learning model along with a feedback block drawing inspiration from [8], [22] by utilizing features extracted from a single LDR image.

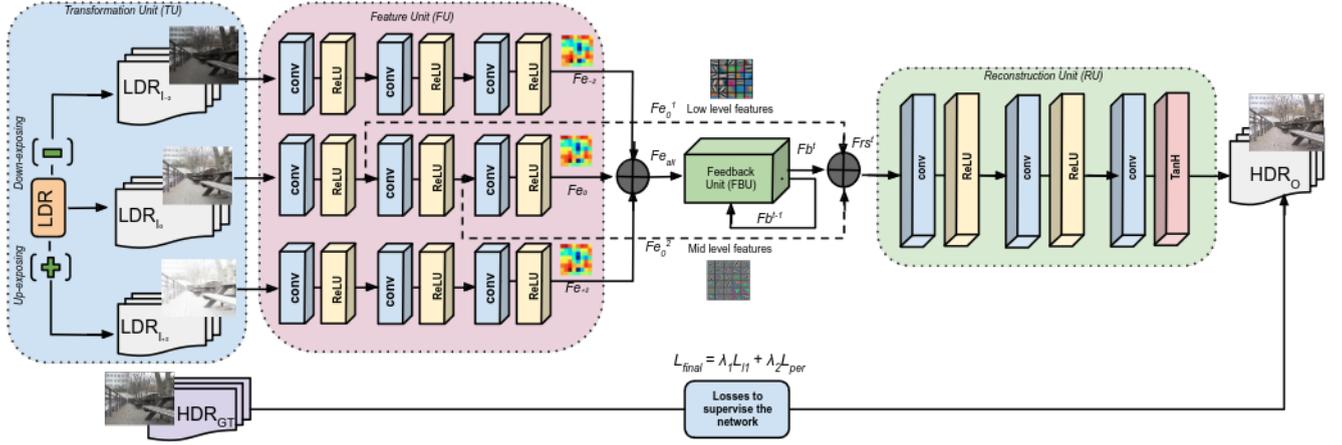

Fig. 2: Illustration of the proposed ArtHDR-Net architecture.

## III. PROPOSED METHOD

This section describes our proposed architecture for HDR image creation using CNN and a feedback mechanism utilizing coarse and fine features extracted from multi-exposed versions of an input LDR image.

### A. Architecture and Model

Our architecture, which we name ArtHDR-Net, comprises four basic units: a) Transformation unit (TU), b) Feature unit (FU), c) Feedback unit (FBU), and d) Reconstruction unit (RU). Figure 2 shows the overall architecture of the method.

TU is responsible for taking an LDR image as input and generating the over-exposed and under-exposed versions (EV ±2). We use an off-the-shelf method [12] to accomplish this task. Regardless of how the input image was captured, we assume that it has a middle exposure value, EV 0. The resulting multi-exposed LDRs ($LDR_{I_0}$, $LDR_{I_{+2}}$, and $LDR_{I_{-2}}$) are used by FU.

FU has three parallel branches for the three LDRs. Each branch has 3×3 kernels in each of the three 2D convolutional layers (conv), followed by rectified linear unit (ReLU) activation function. The three layers (in each branch) help extract three levels of features (i.e., low, mid, and high). We denote the process in FU as $func_{FU}$ and the output of FU (feature information in each branch) as $Fe$ such that:

$$Fe_{-2} = func_{FU}(LDR_{I_{-2}}) \quad (1)$$
$$Fe_0 = func_{FU}(LDR_{I_0}) \quad (2)$$
$$Fe_{+2} = func_{FU}(LDR_{I_{+2}}). \quad (3)$$

The final feature map $Fe_{all}$ is defined as follows:

$$Fe_{all} = Fe_{-2} + Fe_0 + Fe_{+2}. \quad (4)$$

Initially, FBU takes the $Fe_{all}$ as input and subsequently considers the output of FBU from the previous iteration as shown in Figure 2. Similar to [23], [24], we adopt the feedback mechanism with three dilated dense blocks (having dilation rate of 3), which enhances the network's receptive field without increasing the number of parameters and computational time. This choice is made to utilize all the hierarchical features of the input LDRs. FBU also contains two 1 × 1 feature compression layers at the beginning and the end of each of the three dilated blocks. Each of the three dilated blocks contain four 3 × 3 convolutional layers with ReLU activation. This helps to reduce the number of parameters of the network and enhances the learning capability of the network. Finally, FBU starts with a 1 × 1 convolutional layer for compression and ends with a 3 × 3 convolutional layer for further processing. This supports global and local feedback mechanisms in the entire FBU as well as in the three dilated block levels. We denote the process of FBU by $func_{FBU}$ and its output by $Fb$ such that a hidden state defined as:

$$Fb^t = func_{FBU}(Fe_{all}, Fb^{t-1}), \quad (5)$$

where $t$ represents the iteration in the FBU.

When $t = 1$ and $Fb$ is nil, we initialize the hidden state of the FBU with $Fe_{all}$ only. Then we also consider the low-level features, i.e., the coarser (low and mid level) features from the first and second convolutional layers of the middle branch (see Figure 2) of FU which has $LDR_{I_0}$. We do this to guide the network with a set of two residual skip connections for better HDR feature representations in RU. This also helps in preserving the details and artistic intent of the generated HDR image. We denote the final output of the residual skip connections along with the output of FBU at every iteration by $Frs$. We represent this as follows:

$$Frs^t = Fe_0^1 + Fe_0^2 + Fb^t, \quad (6)$$

where $Frs^t$ represents the final feature map (at iteration $t$) after the FBU operates on the feature output. $Fe_0^1$ and $Fe_0^2$ represent the low-level features from $LDR_{I_0}$, which we get as output from the first and second convolutional layers of FU (see Figure 2).

Next, RU reconstruct the HDR image, where it takes the $Frs$ as input in every iteration. RU is made of three $3 \times 3$ 2D convolutional layers with ReLU activation (following the first two layers) and a hyperbolic tangent activation function (TanH) following the third layer (see Figure 2). We denote the function of RU by $func_{RU}$ and the output by $HDR_O$.

$$HDR_O^t = func_{RU}(Frs^t), \quad (7)$$

where $Frs^t$ represents the final learned feature map at the $t^{th}$ iteration and $HDR_O^t$ is the final HDR image generated at iteration $t$.

*B. Loss Functions*

In this work, we calculate the losses between the tone-mapped versions of the generated HDR image ($HDR_{O(TM)}$) and ground truth (GT) HDR image ($HDR_{GT(TM)}$). This is done to avoid the high-intensity values to dominate the loss function values. We use the $\mu - law$ [25] to compress the HDRs. We employ two loss functions, namely, L1 ($L_{l1}$) and perceptual ($L_{per}$) losses, to supervise our model at every iteration $t$. The complete loss function is defined as

$$L_{final} = \lambda_1 L_{l1} + \lambda_2 L_{per}. \quad (8)$$

The $L_{l1}$ loss is the Mean Absolute Error (MAE) calculated between each pixel of $HDR_{GT(TM)}$ and $HDR_{O(TM)}$. The $L_{l1}$ loss is defined as:

$$L_{l1} = \frac{1}{n} \sum_{t=1}^{n} ||HDR_{O(TM)}^t - HDR_{GT(TM)}||. \quad (9)$$

The $L_{per}$ loss is the perceptual loss between $HDR_{GT(TM)}$ and $HDR_{O(TM)}$ generated via the function $func_{per}$ inspired by [26]. $func_{per}$ is a VGG[1] based loss which tries to make the output closer to perceptual similarity. It is created using the ReLU activation layers of the famous 19 layer VGG network. The $L_{per}$ loss is defined as:

$$L_{per} = \frac{1}{n} \sum_{t=1}^{n} func_{per}(HDR_{O(TM)}^t, HDR_{GT(TM)}). \quad (10)$$

We do not consider L2 loss because it is too sensitive towards outliers. We use a combination of the $L_{l1}$ and $L_{per}$ loss because using only the L1 loss produces visually degraded HDRs. $\lambda_1$ and $\lambda_2$ are set to 0.1 and 0.5 empirically.

IV. EXPERIMENTS

We trained and tested our model on Ubuntu 20.04.6 LTS work station with Intel(R) Xeon(R) CPU E5-2687W v3 @ 3.10GHz (20 CPU cores), 126 GB RAM (with 2 GB Swap space) and 1.4 TB SSD for dataset storage. We used batch size of 10 and Adam optimizer [27] for training the model. The model was trained for 150 epochs. The learning rate of the training process was set to $2 \times 10^{-4}$ at first and later adjusted to decay. We employed an iteration count of 4 for the feedback operation in FBU.

*A. Datasets*

We used three datasets for training and testing. Specifically, the City Scene dataset [5] (*ICCV 2017*) contains about 20K HDR-LDR pairs and ground truth HDR images. We also used the LDR-HDR pair dataset from [19] (*IEEE Access 2020*) which contains images from 450 different scenes captured using a Samsung NX3000 camera in three exposure levels (EV -2, 0, +2). This dataset contains images from indoor and outdoor environments as well as scenes containing landscapes, objects, and buildings. It also contains night scenes. Finally we used another dataset, HDR-Synth and HDR-Real from [9] (*CVPR 2020*) which contains real world LDR images and HDR counterparts (9785 in total) as well as synthetic pairs (around 500). We resized the images to $512 \times 512$ before feeding them into the model. We did an 80-20 split on the datasets for training and testing purposes.

*B. Evaluation Metrics*

We use three metrics to evaluate our method quantitatively. For evaluating on the basis of human level perception, we use the *High Dynamic Range Visual Differences Predictor* (HDR-VDP-2 or Q-score) [10], [11]. For structural similarity and accurate reconstruction evaluation, we use the structural similarity index measure (SSIM) and peak signal-to-noise ratio (PSNR). The HDR-VDP-2 score and SSIM are calculated on the actual GT and generated HDRs whereas the PSNR is calculated on the $\mu - law$ based tone-mapped GT and generated HDRs.

V. RESULTS

*A. Quantitative Evaluation*

We evaluate our method ArtHDR-Net against three similar methods [8] (*GlobalSIP 2019*), [9] (*CVPR 2020*), and [12] (*WACV 2023*). We train and test the models on the datasets described in Section IV-A. Table I summarizes the quantitative results in terms of PSNR, SSIM, and HDR-VDP-2 (higher is better). The results confirm that our method is performing well compared to the considered state-of-the-art methods. In the case of City Scene dataset [5], ArtHDR-Net performs the best in terms of PSNR and HDR-VDP-2 scores and second best in terms of SSIM metric. On the other hand, [12] performs best in in terms of SSIM and second best in terms of PSNR scores, but the second best performer in terms of HDR-VDP-2 is [9]. For LDR-HDR pair dataset [19], once again ArtHDR-Net is best in terms of PSNR and HDR-VDP-2 scores. The method in [12] achieves top performance in terms of SSIM metric. Finally for the HDR-Synth and HDR-Real [9] dataset, ArtHDR-Net comes out best in terms of SSIM and HDR-VDP-2 metrics whereas the method in [12] is best in terms of PSNR score. It is noteworthy that our method outperforms all the considered state-of-the-art methods and on all of the considered datasets in terms of HDR-VDP-2 score, which suggests that our method capable of reconstructing HDRs that capture the visual quality of real scenes and better represent all luminance conditions [10], [11]. The methods in [9], [12] are second best in this case. The structural similarity with respect to the input LDRs is better preserved by the method in [12].

---
[1] https://paperswithcode.com/method/vgg-loss

TABLE I: **Quantitative comparison** - PSNR, SSIM and HDR-VDP-2 attained by the proposed method and the state-of-the-art methods. The best results are in **bold**, while the second best results are underlined.

| Method | City Scene dataset [5] | | | LDR-HDR pair dataset [19] | | | HDR-Synth and HDR-Real [9] | | |
|---|---|---|---|---|---|---|---|---|---|
| | PSNR ↑ | SSIM ↑ | HDR-VDP-2 ↑ | PSNR ↑ | SSIM ↑ | HDR-VDP-2 ↑ | PSNR ↑ | SSIM ↑ | HDR-VDP-2 ↑ |
| Method in [8] (2019) | 32.5 | 0.90 | 67 | 15.2 | 0.66 | 64.5 | 17.1 | 0.71 | 66.7 |
| Method in [9] (2020) | 33.4 | 0.91 | <u>68.2</u> | 27.5 | 0.77 | 65.1 | 26 | 0.85 | <u>68</u> |
| Method in [12] (2023) | <u>34</u> | **0.94** | 68 | <u>29</u> | **0.79** | <u>66</u> | **34** | <u>0.87</u> | 67.5 |
| *Ours (ArtHDR-Net)* | **35** | <u>0.93</u> | **69** | **30** | <u>0.78</u> | **66.5** | <u>33.4</u> | **0.88** | **68.3** |

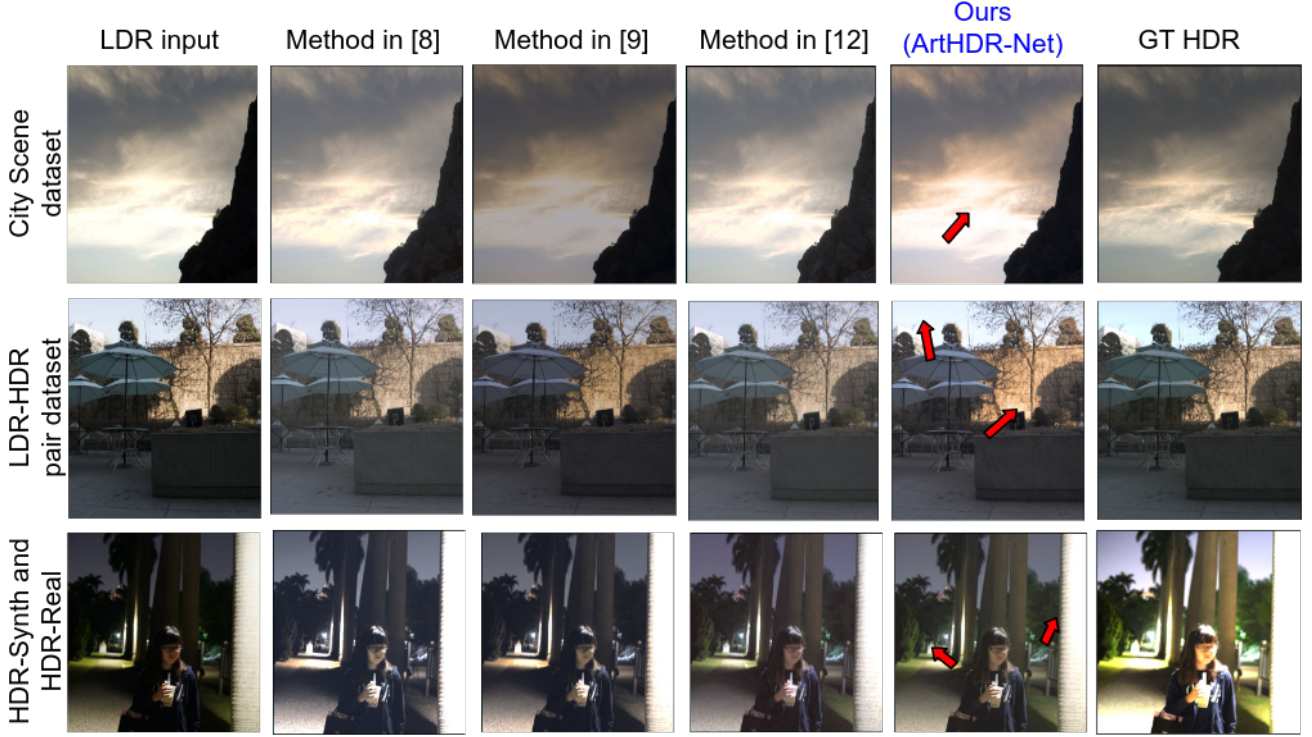

Fig. 3: **Qualitative evaluation** - Output HDR content produced by our method ArtHDR-Net and other state-of-the-art methods for the three datasets. The red arrows in our output highlight the naturalness in color and brightness, hence the artistic intent being preserved with respect to the input LDR image and the GT HDR image.

### B. Qualitative Comparison

Figure 3 depicts the qualitative comparison of our method with the considered conventional methods on the three datasets. The red arrows introduced to the images generated by our method show some sample regions which display naturalness in color and better luminance in comparison to the other methods. The methods in [8], [12] completely fail in this respect while the method in [9] gives similar results to our method, but it is more prone towards dark images. Hence, our method better represents the artistic intent of the images. Similarly, we also evaluate our method on arbitrary images retrieved from the Internet (see Figure 4). In the first row (under-exposed), we see that our method retains good color contrast and luminance compared to the other methods (yellow arrows). The methods in [8], [9] seem closer to ours but still lack in terms of luminance in some regions. In the second row (over-exposed), most methods [8], [12] fail to capture the details and artistic intent of the blue sky where the banding problem is also apparent. In contrast, whereas our method gives more natural and realistic output (red arrows). Although the method in [9] is able to give details of the blue sky to some extent, it still lacks in terms of luminance and color naturalness. The displayed GT as well as the generated HDRs from the various methods (including ours) have been tone-mapped using the algorithm in [28].

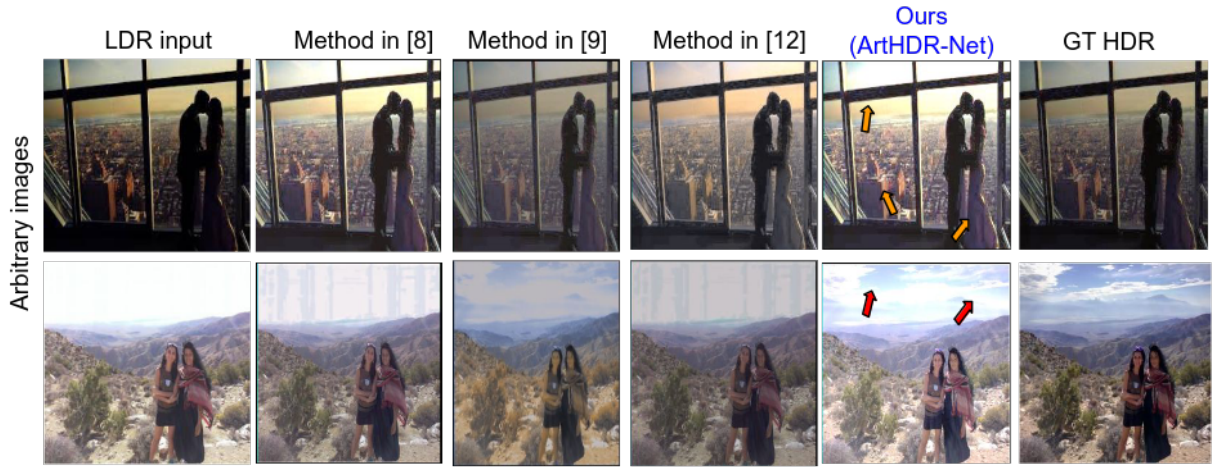

Fig. 4: **Qualitative comparison.** Comparison of our method with the considered state-of-the-art methods with respect to the input LDR and GT HDR (for images not in the mentioned datasets in section IV-A).

TABLE II: **Loss contribution** - The PSNR, SSIM, and HDR-VDP-2 achieved by using the different loss combinations. The best result is in **bold**, while the second best is underlined.

| Loss | PSNR↑ | SSIM↑ | HDR-VDP-2 ↑ |
|---|---|---|---|
| $L_{l1}$ | 36 | 0.931 | 63.5 |
| $L_{per}$ | 35.4 | 0.928 | 66.3 |
| $L_{l1} + L_{per}$ | **36.5** | **0.942** | **67.1** |

TABLE III: **Skip connections contribution** - The PSNR, SSIM, and HDR-VDP-2 achieved by using different combinations of features selected using skip connections to guide the network during the final stage in RU (see Figure 2). The best result is in **bold**, while the second best is underlined.

| $Fe_0^1$ | $Fe_0^2$ | PSNR↑ | SSIM↑ | HDR-VDP-2 ↑ |
|---|---|---|---|---|
| ✗ | ✗ | 35.1 | 0.930 | 63.1 |
| ✗ | ✓ | 35.7 | 0.938 | 64.2 |
| ✓ | ✗ | 35.8 | 0.934 | 64 |
| ✓ | ✓ | **36.5** | **0.942** | **67.1** |

## VI. ABLATION ANALYSIS

We experiment with the components and settings of our model to present the contributions made by of each of them in the achieved output. We create a subset of 5000 arbitrary images (collected from the three datasets described in Section IV-A) to perform the ablation experiments. Table II shows how each of the losses (see Section III-B) contributes to the final accuracy and quality of the HDR images reconstructed using our model. We see that $L_{l1}$ contributes more towards better PSNR and SSIM values, whereas $L_{per}$ have more impact towards the HDR-VDP-2 (Q-score). This suggests that the accuracy of the image reconstruction is mostly influenced by $L_{l1}$, whereas the perceptual quality is maintained by $L_{per}$. Finally, the combination of both losses leads to the best scores for all three metrics. Here, we also weight the losses with $\lambda_1$ and $\lambda_2$ (see Section III-B). Table III illustrates the contribution of each of the skip connections we use in our CNN architecture (see Figure 2). We observe that if we consider the skip connection $Fe_0^1$ only, the PSNR reaches better scores, whereas if we consider only $Fe_0^2$, SSIM and Q-score are better. However, if we consider both together, we get the best results for all three metrics. This implies that more low-level features improve the PSNR, while mid-level features boost up the SSIM as well as Q-score.

## VII. CONCLUSIONS

In this paper, we attempt to improve the state-of-the-art HDR image generator with respect to the artistic quality of the generated HDRs from input LDRs. The quantitative results show the dominance of our method in terms of the HDR-VDP-2 which signifies that our method produces output that is visually more pleasing and appears more realistic than the other considered state-of-the-art methods. The qualitative results also support this argument. In addition, the PSNR and SSIM attained by our method are also comparable to the other state-of-the-art methods.

As future work, we intend to study the impact of multiple EV images (e.g., EV ±1, ±2, ±3 and ±4) on the feature map as well as the behaviour of the model on different EV inputs of the same image. Currently, we do not consider images with foreground motion or dynamic scenes [29]. We aim to upscale our method to videos [1], [20], [21] and dynamic scenes in the future [30]–[32]. Future work will also include the introduction of multiple tasks, such as denoising [33], deblurring, super-resolution [22], [26], and demosaicing into our framework. Finally, we plan to experiment with different loss functions and metrics [14], [15], [34], which will be more efficient in supervising the network in terms of visual quality and preservation of artistic intent.

## ACKNOWLEDGEMENT

This work is supported by Global Excellence and Mobility Scholarship (GEMS)[2], Monash University Malaysia.

---
[2]https://www.monash.edu.my/research/support-and-scholarships/gems-scholarship

## REFERENCES

[1] X. Chen, Z. Zhang, J. S. Ren, L. Tian, Y. Qiao, and C. Dong, "A new journey from sdrtv to hdrtv," in *Proceedings of the IEEE/CVF International Conference on Computer Vision*, 2021, pp. 4500–4509.

[2] N. Matsuda, Y. Zhao, A. Chapiro, C. Smith, and D. Lanman, "Hdr vr," in *ACM SIGGRAPH 2022 Emerging Technologies*, ser. SIGGRAPH '22. New York, NY, USA: Association for Computing Machinery, 2022. [Online]. Available: https://doi.org/10.1145/3532721.3535566

[3] P. Satilmis and T. Bashford-Rogers, "Deep dynamic cloud lighting," *arXiv preprint arXiv:2304.09317*, 2023.

[4] X. Huang, Q. Zhang, Y. Feng, H. Li, X. Wang, and Q. Wang, "Hdr-nerf: High dynamic range neural radiance fields," in *Proceedings of the IEEE/CVF Conference on Computer Vision and Pattern Recognition*, 2022, pp. 18 398–18 408.

[5] J. Zhang and J.-F. Lalonde, "Learning high dynamic range from outdoor panoramas," in *Proceedings of the IEEE International Conference on Computer Vision*, 2017, pp. 4519–4528.

[6] P. Raipurkar, R. Pal, and S. Raman, "Hdr-cgan: single ldr to hdr image translation using conditional gan," in *Proceedings of the Twelfth Indian Conference on Computer Vision, Graphics and Image Processing*, 2021, pp. 1–9.

[7] L. Wang and K.-J. Yoon, "Deep learning for hdr imaging: State-of-the-art and future trends," *IEEE transactions on pattern analysis and machine intelligence*, vol. 44, no. 12, pp. 8874–8895, 2021.

[8] Z. Khan, M. Khanna, and S. Raman, "Fhdr: Hdr image reconstruction from a single ldr image using feedback network," in *2019 IEEE Global Conference on Signal and Information Processing (GlobalSIP)*. IEEE, 2019, pp. 1–5.

[9] Y.-L. Liu, W.-S. Lai, Y.-S. Chen, Y.-L. Kao, M.-H. Yang, Y.-Y. Chuang, and J.-B. Huang, "Single-image hdr reconstruction by learning to reverse the camera pipeline," in *Proceedings of the IEEE/CVF Conference on Computer Vision and Pattern Recognition*, 2020, pp. 1651–1660.

[10] R. Mantiuk, K. J. Kim, A. G. Rempel, and W. Heidrich, "Hdr-vdp-2: A calibrated visual metric for visibility and quality predictions in all luminance conditions," *ACM Transactions on graphics (TOG)*, vol. 30, no. 4, pp. 1–14, 2011.

[11] M. Narwaria, R. K. Mantiuk, M. P. Da Silva, and P. Le Callet, "Hdr-vdp-2.2: a calibrated method for objective quality prediction of high-dynamic range and standard images," *Journal of Electronic Imaging*, vol. 24, no. 1, pp. 010 501–010 501, 2015.

[12] P.-H. Le, Q. Le, R. Nguyen, and B.-S. Hua, "Single-image hdr reconstruction by multi-exposure generation," in *Proceedings of the IEEE/CVF Winter Conference on Applications of Computer Vision*, 2023, pp. 4063–4072.

[13] G. Luzardo, J. Aelterman, H. Luong, W. Philips, D. Ochoa, and S. Rousseaux, "Fully-automatic inverse tone mapping preserving the content creator's artistic intentions," in *2018 Picture Coding Symposium (PCS)*. IEEE, 2018, pp. 199–203.

[14] J. Li and P. Fang, "Hdrnet: Single-image-based hdr reconstruction using channel attention cnn," in *Proceedings of the 2019 4th International Conference on Multimedia Systems and Signal Processing*, 2019, pp. 119–124.

[15] M. S. Santos, T. I. Ren, and N. K. Kalantari, "Single image hdr reconstruction using a cnn with masked features and perceptual loss," *ACM Transactions on Graphics (TOG)*, vol. 39, no. 4, pp. 80–1, 2020.

[16] G. Eilertsen, J. Kronander, G. Denes, R. K. Mantiuk, and J. Unger, "Hdr image reconstruction from a single exposure using deep cnns," *ACM transactions on graphics (TOG)*, vol. 36, no. 6, pp. 1–15, 2017.

[17] G. Luzardo, J. Aelterman, H. Luong, S. Rousseaux, D. Ochoa, and W. Philips, "Fully-automatic inverse tone mapping algorithm based on dynamic mid-level tone mapping," *APSIPA Transactions on Signal and Information Processing*, vol. 9, p. e7, 2020.

[18] G. Cao, F. Zhou, K. Liu, A. Wang, and L. Fan, "A decoupled kernel prediction network guided by soft mask for single image hdr reconstruction," *ACM Transactions on Multimedia Computing, Communications and Applications*, vol. 19, no. 2s, pp. 1–23, 2023.

[19] H. Jang, K. Bang, J. Jang, and D. Hwang, "Dynamic range expansion using cumulative histogram learning for high dynamic range image generation," *IEEE Access*, vol. 8, pp. 38 554–38 567, 2020.

[20] Y. Yang, J. Han, J. Liang, I. Sato, and B. Shi, "Learning event guided high dynamic range video reconstruction," in *Proceedings of the IEEE/CVF Conference on Computer Vision and Pattern Recognition*, 2023, pp. 13 924–13 934.

[21] M. Anand, N. Harilal, C. Kumar, and S. Raman, "Hdrvideo-gan: deep generative hdr video reconstruction," in *Proceedings of the Twelfth Indian Conference on Computer Vision, Graphics and Image Processing*, 2021, pp. 1–9.

[22] Z. Li, J. Yang, Z. Liu, X. Yang, G. Jeon, and W. Wu, "Feedback network for image super-resolution," in *Proceedings of the IEEE/CVF conference on computer vision and pattern recognition*, 2019, pp. 3867–3876.

[23] G. Huang, Z. Liu, L. Van Der Maaten, and K. Q. Weinberger, "Densely connected convolutional networks," in *Proceedings of the IEEE conference on computer vision and pattern recognition*, 2017, pp. 4700–4708.

[24] F. Yu and V. Koltun, "Multi-scale context aggregation by dilated convolutions," *arXiv preprint arXiv:1511.07122*, 2015.

[25] T. Jinno, H. Kaida, X. Xue, N. Adami, and M. Okuda, "$\mu$-law based hdr coding and its error analysis," *IEICE transactions on fundamentals of electronics, communications and computer sciences*, vol. 94, no. 3, pp. 972–978, 2011.

[26] J. Johson, A. Alahi, and L. Fei-Fei, "Perceptual losses for real-time style transfer and super-resolution," Springer, Tech. Rep., 2016.

[27] D. P. Kingma and J. Ba, "Adam: A method for stochastic optimization," *arXiv preprint arXiv:1412.6980*, 2014.

[28] E. Reinhard, M. Stark, P. Shirley, and J. Ferwerda, "Photographic tone reproduction for digital images," in *Seminal Graphics Papers: Pushing the Boundaries, Volume 2*, 2023, pp. 661–670.

[29] Y. Niu, J. Wu, W. Liu, W. Guo, and R. W. Lau, "Hdr-gan: Hdr image reconstruction from multi-exposed ldr images with large motions," *IEEE Transactions on Image Processing*, vol. 30, pp. 3885–3896, 2021.

[30] N. K. Kalantari, R. Ramamoorthi *et al.*, "Deep high dynamic range imaging of dynamic scenes." *ACM Trans. Graph.*, vol. 36, no. 4, pp. 144–1, 2017.

[31] S. Wu, J. Xu, Y.-W. Tai, and C.-K. Tang, "Deep high dynamic range imaging with large foreground motions," in *Proceedings of the European Conference on Computer Vision (ECCV)*, 2018, pp. 117–132.

[32] Z. Pu, P. Guo, M. S. Asif, and Z. Ma, "Robust high dynamic range (hdr) imaging with complex motion and parallax," in *Proceedings of the Asian Conference on Computer Vision*, 2020.

[33] B. Mildenhall, P. Hedman, R. Martin-Brualla, P. P. Srinivasan, and J. T. Barron, "Nerf in the dark: High dynamic range view synthesis from noisy raw images," in *Proceedings of the IEEE/CVF Conference on Computer Vision and Pattern Recognition*, 2022, pp. 16 190–16 199.

[34] H. Zhao, O. Gallo, I. Frosio, and J. Kautz, "Loss functions for image restoration with neural networks," *IEEE Transactions on computational imaging*, vol. 3, no. 1, pp. 47–57, 2016.